\documentclass[journal]{IEEEtran}
\usepackage[T1]{fontenc}
\usepackage{graphicx}
\usepackage{times}
\usepackage{helvet}
\usepackage{courier}
\usepackage{amsmath}
\usepackage{algorithm}
\usepackage{algorithmic}
\usepackage{csquotes} 
\usepackage{color}
\usepackage{paralist}
\usepackage{amssymb}
\usepackage{indentfirst}
\usepackage{subfigure}
\usepackage{float}
\usepackage{multirow}
\usepackage{cite}
\usepackage{mathrsfs}

\usepackage{mathrsfs}
\usepackage{graphicx}
\usepackage{amsmath}
\usepackage{amssymb}
\usepackage{booktabs}
\usepackage{pifont}
\newcommand{\cmark}{\ding{51}}%
\newcommand{\xmark}{\ding{55}}%
\usepackage{makecell}
\usepackage{colortbl}
\definecolor{mygray}{gray}{.9}
\definecolor{mypink}{rgb}{.99,.91,.95}
\definecolor{mycyan}{cmyk}{.3,0,0,0} 

\usepackage{xcolor}
\definecolor{citecolor}{HTML}{0071bc} 
\definecolor{SeaGreen4}{RGB}{0,205,102} 
\definecolor{SlateBlue}{RGB}{106,90,205} 
\definecolor{DarkRed}{RGB}{178,34,34}

\usepackage{threeparttable}
\usepackage{multirow}
\usepackage{longtable}
\usepackage{rotating}
\usepackage{tabularx}
\usepackage{color}
\usepackage{xcolor}
\definecolor{citecolor}{HTML}{0071bc}
\usepackage[colorlinks, linkcolor=red,  anchorcolor=blue, citecolor=citecolor]{hyperref}

\usepackage[misc]{ifsym}

\usepackage{textcomp,booktabs}
\usepackage{amssymb}
\usepackage{pifont}

\usepackage{colortbl}
\definecolor{mygray}{gray}{.9}
\definecolor{mypink}{rgb}{.99,.91,.95}
\definecolor{mycyan}{cmyk}{.3,0,0,0}

\begin{document}

\title{VisEvent: Reliable Object Tracking via Collaboration of Frame and Event Flows} 

\author{Xiao Wang, \emph{Member, IEEE}, Jianing Li, Lin Zhu, Zhipeng Zhang, Zhe Chen, \emph{Member, IEEE},  Xin Li, \\ 
Yaowei Wang, \emph{Member, IEEE}, Yonghong Tian, \emph{Fellow, IEEE}, Feng Wu, \emph{Fellow, IEEE} 
\thanks{
Xiao Wang is with School of Computer Science and Technology, Anhui University, Hefei 230601, China. He is also with Peng Cheng Laboratory, Shenzhen, China. (email: xiaowang@ahu.edu.cn) 

Xin Li, and Yaowei Wang are with Peng Cheng Laboratory, Shenzhen, China. (email: xinlihitsc@gmail.cn, wangyw@pcl.ac.cn) 

Lin Zhu, Jianing Li, Yonghong Tian are with Peng Cheng Laboratory, Shenzhen, China, and National Engineering Laboratory for Video Technology, School of Electronics Engineering and Computer Science, Peking University, Beijing, China. (email: \{lijianing, linzhu, yhtian\}@pku.edu.cn) 

Zhipeng Zhang is with National Laboratory of Pattern Recognition, Institute of Automation, Chinese Academy of Sciences, and School of AI, University of Chinese Academy of Sciences. (email: zhangzhipeng2017@ia.ac.cn)  

Zhe Chen is with The University of Sydney, Australia. He is also with the Cisco-La Trobe Centre for Artificial Intelligence and Internet of Things, La Trobe University. Address: Edwards Rd, Flora Hill 3552, Australia. (email: zhe.chen@latrobe.edu.au)

Feng Wu is with the University of Science and Technology of China, Hefei, China. (email: fengwu@ustc.edu.cn) 

\Letter~~Corresponding author: Yaowei Wang  
}}

\markboth{IEEE Transactions on Cybernetics} 
{Shell \MakeLowercase{\textit{et al.}}: Bare Demo of IEEEtran.cls for IEEE Journals}

\maketitle

\begin{abstract}
Different from visible cameras which record intensity images frame by frame, the biologically inspired event camera produces a stream of asynchronous and sparse events with much lower latency. In practice, visible cameras can better perceive texture details and slow motion, while event cameras can be free from motion blurs and have a larger dynamic range which enables them to work well under fast motion and low illumination.  Therefore, the two sensors can cooperate with each other to achieve more reliable object tracking. In this work, we propose a large-scale Visible-Event benchmark (termed VisEvent) due to the lack of a realistic and scaled dataset for this task. Our dataset consists of 820 video pairs captured under low illumination, high speed, and background clutter scenarios, and it is divided into a training and a testing subset, each of which contains 500 and 320 videos, respectively. Based on VisEvent, we transform the event flows into event images and construct more than 30 baseline methods by extending current single-modality trackers into dual-modality versions. More importantly, we further build a simple but effective tracking algorithm by proposing a cross-modality transformer, to achieve more effective feature fusion between visible and event data. Extensive experiments on the proposed VisEvent dataset, FE108, COESOT, and two simulated datasets (i.e., OTB-DVS and VOT-DVS), validated the effectiveness of our model. The dataset and source code have been released on: \url{https://github.com/wangxiao5791509/VisEvent_SOT_Benchmark}. 
\end{abstract}

\begin{IEEEkeywords}
Visual Tracking; Neuromorphic Vision; Dynamic Vision Sensors; Event Camera; Self-attention and Transformers. 
\end{IEEEkeywords}

\IEEEpeerreviewmaketitle

\section{Introduction} 
\IEEEPARstart{V}{isual} tracking aims at locating the initialized object in the first frame with a bounding box and adjusting the box to better fit the target object for subsequent video frames. It has been widely used in many applications, including intelligent video surveillance, robotics, and autonomous vehicles. 
With the help of deep learning, many representative deep trackers are proposed~\cite{zhang2021learn, wang2021towards, wang2021gantrack, wang2021deepmta, dong2020clnet, shen2021distilled, dong2018tripletSiam, han2021learning, wang2022beamTrack}. To be specific, Dong et al.~\cite{dong2020clnet} find that the samples given in the first frame (termed decisive samples) are ignored during offline training and propose a compact latent network that can quickly adjust the tracking model to work well in new scenarios. Shen et al.~\cite{shen2021distilled} propose a teacher-student knowledge distillation model that supports the learning of a small but fast tracker from large Siamese trackers. Dong et al.~\cite{dong2018tripletSiam} also exploit a new triplet loss function to enhance the deep features for Siamese-based tracking. Different from existing Siamese trackers which adopt depth-wise cross-correlation (DW-XCorr) for activation response maps prediction, Han et al.~\cite{han2021learning} propose the asymmetric convolution (ACM) operators for tracking which is a learnable and flexible module. However, due to the utilization of RGB cameras, existing trackers still suffer from challenging scenarios such as~\emph{low illumination}, \emph{fast motion}, and \emph{background clutter}.

\begin{figure}
\center
\includegraphics[width=3.5in]{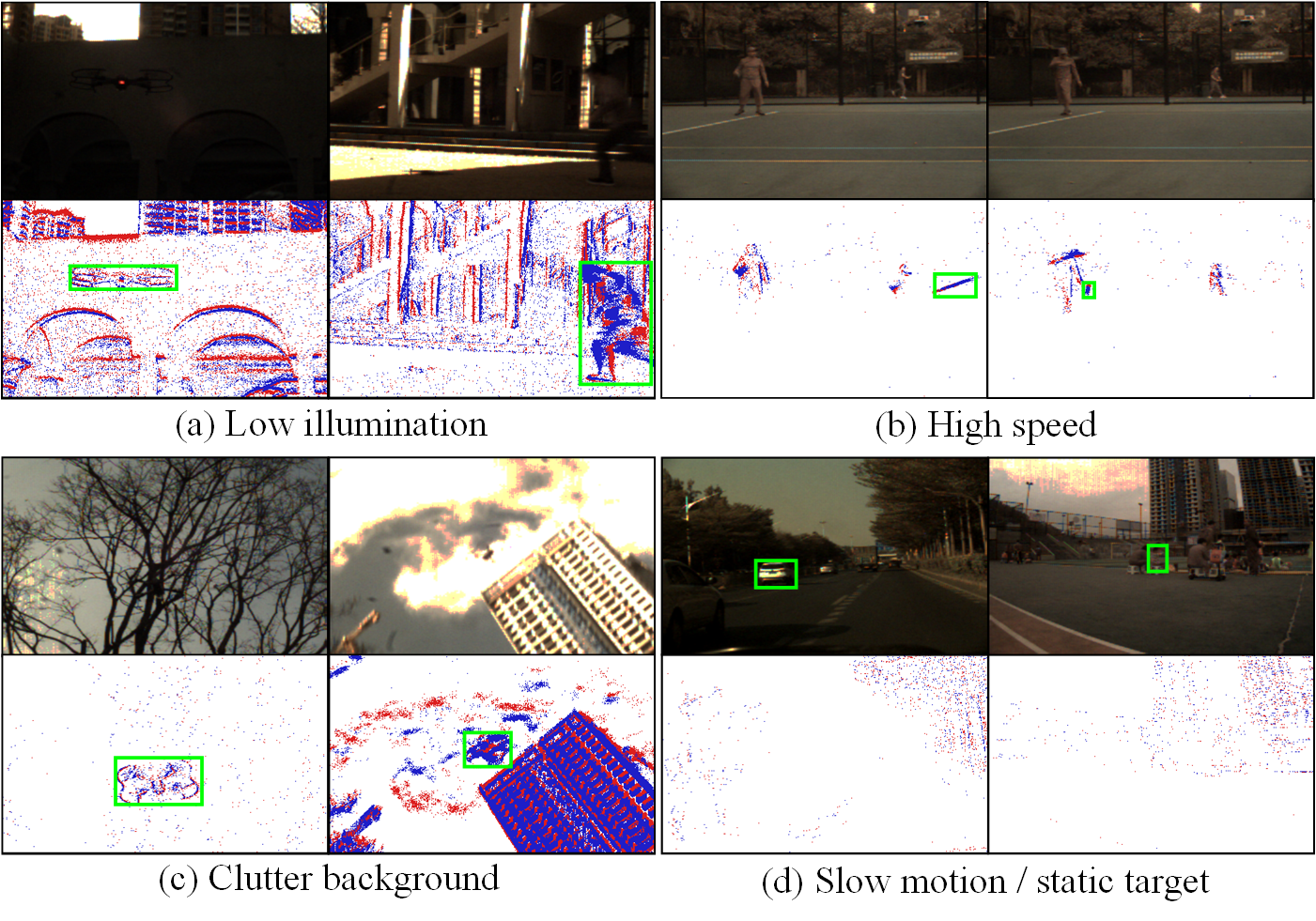}
\caption{
\textcolor{black}{Illustration of complementary characteristics of Visible and Event cameras. From subfigure (a-c), we can find that the Event camera works well in low illumination, high speed, and even in clutter background scenarios due to its advantages in high dynamic range, dense temporal resolution, and unique imaging features. In contrast, the Visible camera performs poorly when facing these challenging attributes, however, it works well in slow motion or static scenarios and is good at capturing the color and texture information according to subfigure (d). Therefore, a more reliable and accurate tracking result can be obtained if we combine the two sensors.}
} 
\label{frontimg}
\end{figure}

\begin{figure} 
\center
\includegraphics[width=3in]{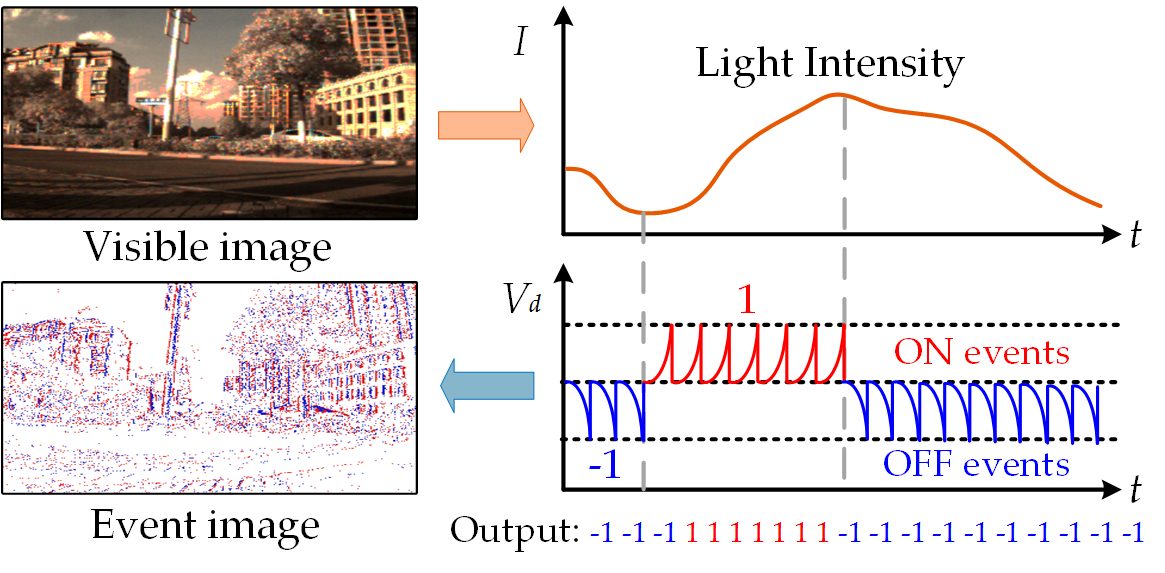}
\caption{Sampling mechanisms of visible and event cameras. Each pixel in the visible image records the light intensity (I) in a synchronous way; while each pixel in the biologically inspired event camera asynchronously reflects the changes in lighting intensity. Usually, we use 1/-1 to denote the ON/OFF event (enhancement/diminished light). $V_d$ is the neuronal membrane potential.}
\label{sampleScheme}
\end{figure}

To handle aforementioned challenges, some researchers resort to biologically inspired event cameras like Dynamic Vision Sensors (DVS) \cite{2008dvs} for target object tracking \cite{chen2020eventTrack, ramesh2020etld, camunas2017event, chen2019asynchronouseventtrack, yang2019dashnet}. Different from regular visible cameras which record an intensity image in a \emph{frame} manner (high latency, i.e., 10-20 ms), the event cameras output a \emph{stream of asynchronous events}. The pixels of event cameras send information independently only when visual intensity changes (also called an event). Therefore, the event sensors excel at capturing the motion information with very low latency (1 $\mu$s) and are almost free from the trouble of motion blur. It also requires much less energy, bandwidth, and computation. In addition, DVS sensors also outperform the visible cameras on dynamic range (140 vs 60dB), which enables them to work effectively even in poor illumination conditions. The advantages of the latency, resource consumption, and operation environments make the event cameras more suitable for target tracking in challenging scenarios. The comparison of the imaging quality and sampling mechanism of the two sensors are given in Fig. \ref{frontimg} and Fig. \ref{sampleScheme} respectively, to help readers better understand their unique advantages.

Despite benefits, the event cameras can't capture slow-motion or static objects and lack fine-grained texture information which is also very important for high-performance tracking. Therefore, the integration of visible cameras and DVS sensors is an intuitive idea for reliable object tracking. There are already several works \cite{liu2016combined, chen2019asynchronouseventtrack, huang2018event, yang2019dashnet} developed based on this setting, but their experiments are conducted on simulation data or several simple videos. Their performance on real data in the wild is still unknown. Recently, Zhang et al. proposed a new dataset that contains 108 videos, termed FE108 \cite{Zhang2021FE108}, but tracking performance on this dataset is almost saturated. The development of this field is rather slow compared with visible camera based tracking due to the lack of a large-scale Visible-Event based object tracking dataset.

In this work, we first propose a large-scale neuromorphic tracking benchmark that contains 820 Visible and Event video sequence pairs, termed VisEvent. This dataset fully reflects the challenging factors in the real world like motion blur, fast and slow motion, low illumination, high dynamic range, background clutter, etc. It contains 17 attributes and mainly focuses on traffic scenes, thus the target objects are mainly people and vehicles. To construct a comprehensive benchmark, we also recorded some videos from the indoor scene. In total, our dataset contains $371,128$ frames. We split them into the training and testing subsets, each of them containing 500 and 320 video pairs respectively. Due to the lack of baseline methods to be compared for future works, we extend currently visible camera based trackers into dual-modality versions with different fusion strategies like early, middle, and late fusion.

On the basis of our newly proposed VisEvent dataset, we further build a novel and effective baseline method by developing a Cross-Modality Transformer module, termed CMT. 
The proposal of our CMT module is based on the following two observations and key insights: 
1). Existing trackers~\cite{Nam2015Learning, Jung_2018_ECCV, danelljan2019atom} usually adopt convolutional neural networks for tracking, which only learn the local features well using limited convolutional kernels. The most recent RGB trackers also exploit self-attention or Transformers networks for global representation learning~\cite{yan2022unificationTrack, chen2021TransT}. 
2). Different from existing Transformer based trackers, we need to consider the modality interactions between RGB and Event streams to achieve a more robust feature fusion. Therefore, a cross-attention layer is proposed to connect the dual modalities for interactive message passing. Then, a self-attention layer is adopted to learn and enhance the feature representations in a global view. Thus, our proposed CMT boosts the inter- and intra-modality features and significantly improves the final tracking results. As validated in our experiments, it can be integrated into existing binary classification trackers like MDNet~\cite{Nam2015Learning}, RT-MDNet~\cite{Jung_2018_ECCV}, and discriminative correlation filter trackers ATOM~\cite{danelljan2019atom}. More details can be found in Fig.~\ref{pipeline} and Section~\ref{CMT}.

Generally speaking, the contributions of this paper can be concluded as the following three aspects \footnote{The video tutorial of this work can be found at \url{https://youtu.be/vGwHI2d2AX0}}:

$\bullet$ \textcolor{black}{We introduce a comprehensive neuromorphic tracking dataset comprising 820 Visible-Event videos, termed VisEvent. This marks the inception of a large-scale Visible-Event benchmark dataset collected from real-world scenarios, tailored specifically for single object tracking.}

$\bullet$ \textcolor{black}{We present a straightforward yet highly effective baseline tracker achieved through the development of a cross-modality transformer module. This module adeptly leverages the distinctive characteristics of various modalities to enhance tracking robustness. Notably, this is the first instance where the successful application of a cross-modality transformer in visible-event tracking has been demonstrated.} 

$\bullet$ \textcolor{black}{We have assembled a diverse set of more than 35 dual-modality-based trackers for our benchmark dataset. These trackers serve as valuable resources for future research, enabling comprehensive comparisons across various tracking pipelines (e.g., correlation filter-based, binary classification-based, and Siamese matching-based trackers) and fusion strategies (e.g., early, middle, and late fusion).}

\section{Related Work} 

\noindent 
\textbf{Visible Camera based Tracking. }
Most of the current trackers are developed based on RGB cameras and track the target object frame by frame. Traditional RGB trackers use hand-crafted features for target representation but perform poorly in challenging scenarios. Among them, correlation filter (CF) based trackers dominate the tracking field due to their high efficiency and good performance~\cite{han2019STCATrack, han2019stateDriftTrack, han2017adaTrack}. After that, the deep learning trackers, especially the Siamese network based trackers began to occupy the top positions of various benchmarks. Shen et al. \cite{shen2019trackSiamHier} propose an attention based Siamese network which can improve matching performance by a sub-Siamese network. The feature learning and classification capabilities of extreme learning machine (ELM) is exploited by Deng et al. \cite{Deng2020ELMF} for efficient visual tracking. Liu et al. \cite{Liu2020FSMs} attempt to address the occlusion issue in the tracking task using correlation filtering and probabilistic finite state machines (FSMs). Zhou et al. \cite{Zhou2022TargetTrack} propose a gradient-guided feature adjustment module to generate target-aware features for constructing the state estimation network, which achieves high-performance visual tracking. Li et al. \cite{li2020dualregressTrack} develop a dual-regression tracking framework by combining the discriminative fully convolutional module and a fine-grained correlation filter component. Li et al. \cite{li2019hierarchicalSASiam} treat the TIR tracking as a similarity verification task, and propose a Hierarchical Spatial-aware Siamese CNN (named HSSNet) for TIR tracking. Dong et al.~\cite{dong2019hyperparamdrl} exploit the parameter optimization in tracking task using deep reinforcement learning and achieve a higher tracking performance. Lu et al.~\cite{lu2022deepShrinkageloss} propose a new shrinkage loss to handle the data imbalance issue when learning deep features for tracking. Dong et al.~\cite{dong2019quadruplet} argue that the training instances are ignored in existing trackers and the authors propose a new quadruplet deep network that obtains more powerful features for single object tracking. Liang et al.~\cite{liang2019localsemanticSiam} propose the Local Semantic Siamese (LSSiam) network which not only learns global features well but also local semantic features (which contains more fine-grained and partial information) for visual tracking. 
\textcolor{black}{
Cao et al.~\cite{cao2023trackcontrol} propose an autonomous underwater vehicle tracking control algorithm that handles the underwater dynamic target tracking task well by predicting its trajectory. 
Choi et al.~\cite{choi2023SphereTracking} address the task of target tracking on the sphere by considering topographic structure. 
}

\textcolor{black}{
Recently, the Transformer networks are widely exploited in visual tracking task and achieve higher performance on multiple benchmark datasets~\cite{chen2021TransT}.   
}
However, their performance under low illumination, fast motion, and low resolution is still unsatisfactory. Many works are proposed to handle these issues like active hard sample generation \cite{Wang_2018_CVPR, song2018vital} and deblur \cite{guo2019effects}, however, the existing algorithm can't address these issues well due to the bad imaging quality of RGB cameras. Other sensors are also explored for tracking task, including high frame rate cameras (short for HFR, larger than 200 FPS), thermal cameras, and depth cameras, but HFR cameras are sensitive to illumination, thermal cameras are expensive, and depth cameras are also helpless for high speed and low light, which limit their wide applications in practical scenarios.

\noindent 
\textbf{Event Camera based Tracking. }
Compared with RGB trackers, few people pay attention to tracking based on event cameras. Chen et al. \cite{chen2019asynchronouseventtrack, chen2020eventTrack} propose the Adaptive Time-Surface with Linear Time Decay (ATSLTD) event-to-frame conversion algorithm for event frame construction and re-detect the target object when model drifting. The synchronous Time-Surface with Linear Time Decay (TSLTD) representation is explored and fed into a CNN-LSTM network for 5-DoF object motion regression in \cite{chen2020eventTrack}. To handle the issue of local search in event based tracking, the authors of \cite{ramesh2018eventlong} propose a data-driven, global sliding window based detector to help re-detect the target object when it re-enters the field-of-view of the camera. \cite{chamorro2020highevent, alzugaray2020haste} explore the high-speed feature tracking with DVS sensors. Cao et al. \cite{cao2015SNNtrack} propose a target tracking controller based on a spiking neural network that can be deployed on autonomous robots. Jiang et al. \cite{jiang2020object} also propose a tracking framework that contains an offline-trained detector and an online-trained tracker which complement each other. 
Zhu et al.~\cite{zhu2022GrapheventTrack} propose a density-insensitive downsampling strategy to get the key events and employ graph neural networks to capture the spatiotemporal cues for event based tracking.
Zhang et al.~\cite{zhang2023fealignTrack} exploit the cross-style and cross-frame-rate alignment between the visual and event data for accurate tracking. 
Although these trackers work well in simple scenarios, however, their performance on large-scale tracking benchmarks is still unknown. Also, the performance of these models on tracking objects that rarely move or are stationary is still alarming.

\noindent 
\textbf{Tracking by Combining Visible and Event Cameras.} 
Joint utilizing the two sensors for robust tracking is an intuitive idea and the initial verification has been obtained in the following work. For example, \cite{gehrig2018asynchronous, gehrig2020eklt} first propose asynchronous photometric feature tracking with the event and RGB sensors. Liu et al. \cite{liu2016combined} also attempt to extract candidate ROIs from RGB frames and event flows simultaneously for more accurate tracking. Huang et al. \cite{huang2018event} develop an SVM-based tracker using re-constructed samples for an online update and candidate search locations mining from event flows with a CeleX sensor. DashNet \cite{yang2019dashnet, zhao2022HNNframework} is developed based on parallel SNN and CNN tracking and fusion which can run at 2083 FPS on neuromorphic chips. Their work fully demonstrates the vast potential of Visible-Event tracking in practical applications. \textcolor{black}{Tang et al.~\cite{tang2022coesot} propose a single-stream multi-modal tracking framework based on the Transformer network which directly takes the RGB frame and event voxel as input.}

These works have made preliminary explorations in this direction, however, their experiments are conducted on several simple real videos or simulation data, as shown in Table \ref{benchmarkList}. Their results on really challenging scenarios are still unknown, also, their work lacks proper baseline methods to compare. We believe our proposed dataset and baseline algorithm will be a good platform for research in this direction.

\section{Methodology}

\subsection{Motivation and Overview} \label{motivationView}
Based on our proposed VisEvent dataset, we first extend current trackers which are developed for RGB videos into dual-modality versions and evaluate their results. According to the experimental results, we observe that existing trackers are less effective on our dataset even though deep neural networks and regular attention modules are used. For instance, as shown in Fig. \ref{benchmarkResults}, the SiamRPN++ \cite{li2018siamrpn++} and SuperDiMP \cite{bhat2019DiMP} only attains $0.576|0.410$ and $0.489|0.320$ on precision and success plot, respectively. MDNet \cite{Nam2015Learning} with channel and spatial attention only achieves $0.456|0.273$ and $0.455|0.270$, as listed in Table \ref{FusionAnalysis}. How to design a more effective information fusion module for visible-event tracking is still a question worth exploring.

\begin{figure*}
\center
\includegraphics[width=7in]{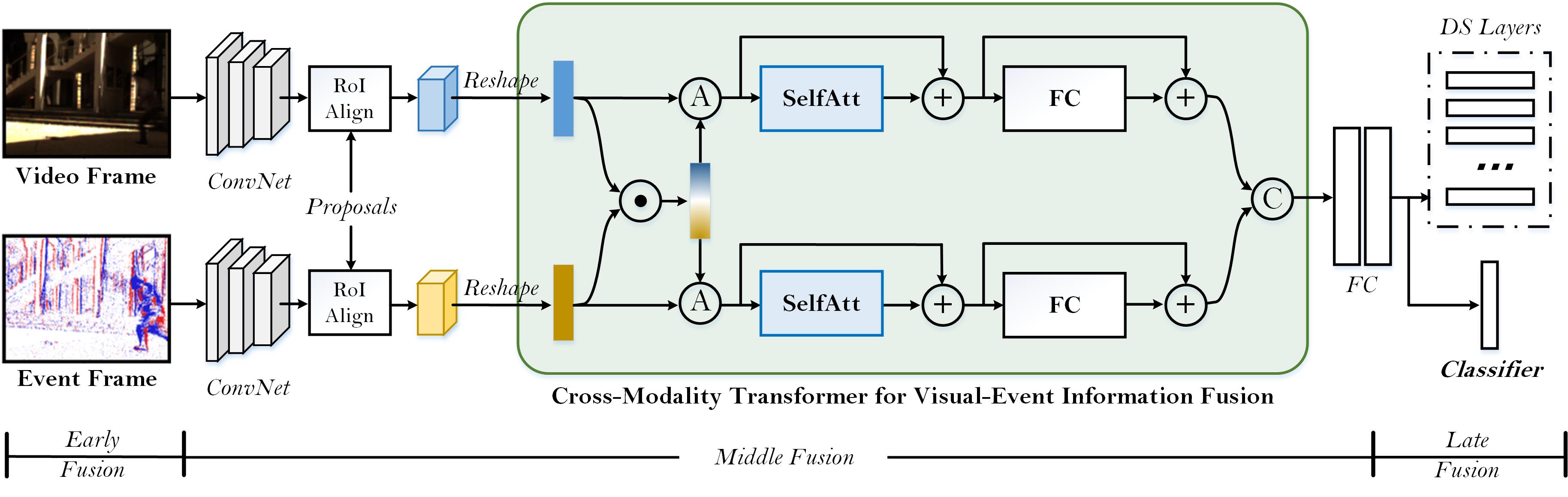}
\caption{ 
\textcolor{black}{An overview of our proposed tracking framework via collaboration of visible frame and event streams. The RT-MDNet tracker is adopted as an example to demonstrate our tracking procedure. Given the RGB and Event frames, we first extract the positive and negative training samples from the first frame to learn a classifier. Three convolutional layers are used to extract the deep feature maps. Then, the RoI Align operator is adopted to extract the instance-level features given the extracted proposals for both modalities. The RGB and Event features are first connected using the dot product to get the base vector, then, the cross-attention is conducted for each modality to enhance the message passing. Self-attention is proposed to learn the global features that are complementary to local CNN features. Finally, we feed the feature vectors into fully connected layers for proposal classification. The best-scored proposal is selected as the tracking result of the current step and similar procedures are repeated until the end of the testing video.} }     
\label{pipeline}
\end{figure*}

\textcolor{black}{
In this paper, we propose a new tracking algorithm by fully exploiting the RGB frame and event stream. The key motivation of our tracker is existing trackers usually adopt convolutional neural networks for tracking, which only learn the local features well using limited convolutional kernels. The most recent RGB/Event trackers also exploit self-attention or Transformer networks for global representation learning~\cite{zhang2022STNet, chen2021TransT}. On the other hand, the modality interactions between RGB and event stream need to be considered for high-performance tracking. Given the input modalities, 
}
we simply use the event images transformed from event flows to fuse with visible images. The convolutional neural network is used for feature extraction, more importantly, a simple but effective feature fusion module is proposed to achieve interactive learning via information propagation between dual modalities, termed cross-modality transformers (CMT). As shown in Fig. \ref{pipeline}, we first extract the feature representations of visible and event images using CNN. Then, the candidate proposals are sampled and fed into the RoI align module for instance feature extraction. Afterward, we attain the base vector $m$ with element-wise multiplication operations based on the given feature representations of dual modalities. The base vector is used as the query vector to attend the two modalities (context vector) respectively to realize the interaction and transmission of information flows. After that, we use self-attention layers to boost the internal connections of each modality. Lastly, the two features are concatenated and fed into the classifier for tracking.

\subsection{Input Representation}  \label{InputRepresent} 

Given the synchronous video frames and asynchronous event flows, how to represent and adaptively fuse the two modalities is the key to successful and reliable visual tracking on the VisEvent. From the perspective of the perception principle, the visible cameras capture the global scene by recording the intensity of all pixels in a \emph{frame} manner. Usually, the CNN is used to extract its feature representations, for example, the RT-MDNet \cite{Jung_2018_ECCV} uses three convolutional layers as its backbone network. Different from visible sensors, the event cameras asynchronously capture the variation in log-scale intensity ($I$), in other words, each pixel will output a discrete event independently when the visual change exceeds a threshold ($\theta$): 
\begin{equation}
\small 
||log(I_{t+1}) - log(I_t)|| \geq \theta
\end{equation}
In practice, we use a 4-tuple $\{x, y, t, p\}$ to represent the discrete event of a pixel captured with DVS, where the $x, y$ are spatial coordinates, $t$ is the timestamp, and $p$ is the polarity of brightness variation, i.e., 1 and -1 are used to denote the ON event (increase) and OFF event (decrease) respectively. A comparison of the sampling mechanism of visible and event cameras is visualized in Fig. \ref{sampleScheme}.

To fully utilize the benefits of CNN, previous event trackers \cite{ramesh2018eventlong, chen2019asynchronouseventtrack, chen2020eventTrack} usually transform the asynchronous event flows into synchronous \emph{event image} by stacking the events in a fixed time interval. In this work, we also adopt such transformation to get the event images but focus on designing novel feature fusion modules for high-performance tracking. In the subsequent subsection, we will introduce our proposed cross-modality transformer for interactive dual-modal information propagation.

\subsection{Cross-Modality Transformer for Fusion}   \label{CMT}
Following RT-MDNet \cite{Jung_2018_ECCV}, we take the three convolutional layers as the shared backbone of our tracker. Once we obtain the feature representation of dual modalities, we can directly fuse them for tracking. However, the dual features are extracted independently and thus lack interactive feature learning which may limit its representation ability. Many works demonstrate that joint feature learning between multi-modal data will bring more powerful feature representation. Inspired by previous works \cite{vaswani2017attention, tan2019lxmert}, the Cross-Modality Transformer (termed CMT) is proposed in this work to enhance the message passing between dual modalities. This module is developed based on an attention mechanism that targets at retrieving information from \emph{context vectors} $y_j$ based on \emph{query} $x$. Usually, we can first compute the similarity score $a_j$ between the query $x$ and context vector $y_j$ using MLP layers. Then, this score will be normalized with the Softmax operator. Finally, the context vectors will be weighted and summed as the output of an attention layer: $Att_{X \rightarrow Y} (x, \{y_j\}) = \sum_{j} \alpha_j y_j$. The widely used \emph{self-attention layer} \cite{wang2018nonlocal} is a special case of attention family, as its query vector $x$ is actually from the context vectors.

In our scenario, we have dual-modality features that can be used to attend to each other using a cross-attention model, i.e., from RGB to event, and from event to RGB. As shown in Fig. \ref{pipeline}, the features of RGB frame and event flows are firstly fed into a cross-attention model to guide the information propagation along both directions. Formally, we use $F_{v}$ and $F_{e}$ to denote the initial feature obtained from the CNN backbone network. Then, these two feature maps are added along the channel dimension and reshaped into feature vectors $\bar{F_{v}}$ and $\bar{F_{e}}$. A base vector $m$ can be attained by element-wise product between input features of dual-modalities, i.e., $m = \bar{F_{v}} \odot \bar{F_{e}}$. The base vector $m$ is used as the query feature to attend the context vectors, i.e., visible and event features respectively: 
\begin{equation}
\label{CrossAtt}
\small 
\tilde{F_e} = CrossAtt_{V \rightarrow E} (m, F_e), 	  \tilde{F_v} = CrossAtt_{E \rightarrow V} (m, F_v)
\end{equation}
Therefore, the joint cross-modality representations can be attained with a cross-attention model which can align the dual modalities by exchanging the information.

To boost the internal connections, we introduce the self-attention layers \cite{wang2018nonlocal} based on the output of cross-attention model: 
\begin{equation}
\label{SelfAtt1}
\small 
\mathcal{F}_e = SelfAtt_{E \rightarrow E} (\tilde{F_e}, \tilde{F_e}),    \mathcal{F}_v = SelfAtt_{V \rightarrow V} (\tilde{F_v}, \tilde{F_v})  
\end{equation}
For simplicity, we take the event feature $\tilde{F_e}$ as an example, and similar operations are implemented for visible features. Specifically speaking, we first use two FC layers with weights $W_h$ and $W_g$ to process the input $\tilde{F_e}$ separately. The output will be fed into a Softmax layer and multiplied with the results of another branch (i.e., $W_o * \tilde{F_e}$). Finally, we feed these results into an FC layer with weights $W_p$ to get the attended event features. The aforementioned process can be summarized as: 
\begin{equation}
\label{SelfAtt2}
\small 
\mathcal{F}_e = \textbf{$W_p$}(Softmax(   (\textbf{$W_h$} * \tilde{F_e}) * (\textbf{$W_g$} * \tilde{F_e})) * (\textbf{$W_o$} * \tilde{F_e}))
\end{equation}
where $W_p, W_h, W_g$ and $W_o$ are weights of different FC layers, $*$ is the multiplication operation. Then, we feed the attended features into the FC layers to output the final feature vectors.

\subsection{Training and Tracking Phase}  \label{lossfunctions}
\textcolor{black}{
In this work, we follow a binary classification-based tracking framework \cite{Jung_2018_ECCV} and conduct tracking by discriminating whether the given proposal is a target object or not. In the training phase, we introduce a set of domain-specific layers (DS layers) for each video sequence to learn the shared features that are only used in the training phase. The \emph{binary cross-entropy loss} $\mathcal{L}_{ce}$ and \emph{instance embedding loss} $\mathcal{L}_{ie}$ are used for the optimization of our network for the RT-MDNet, i.e., $\mathcal{L} = \mathcal{L}_{ce} + \mathcal{L}_{ie}$. 
To be specific, the $\mathcal{L}_{ce}$ can be formulated as: 
\begin{equation} 
\label{LclsFunction}
\mathcal{L}_{ce} = -\frac{1}{N} \sum_{j=1}^{N}\sum_{c=1}^{2} [\textbf{y}_j]_{c\hat{d}(b)} \cdot log([\sigma_{cls} (\textbf{f}_j^{\hat{d}(b)})]_{c\hat{d}(b)})  
\end{equation} 
where $\textbf{y}_j$ is the ground truth. The value of $[\textbf{y}_j]_{cd}$ is one when the class of a bounding box in the domain $d$ is $c$. $b$ denotes the iterate index. The predicted score of $j$-th proposal is $\textbf{f}_j$. 
For the \emph{instance embedding loss} $\mathcal{L}_{ie}$, 
\begin{equation}
\label{LinstFunction}
\mathcal{L}_{ie} = -\frac{1}{N} \sum_{j=1}^{N}\sum_{d=1}^{D} [\textbf{y}_j]_{+d} \cdot log([\sigma_{inst} (\textbf{f}_j^{d}]_{+d}) 
\end{equation}
Here, the number of domains is $D$. We refer the readers to check their paper for the details of the two loss functions \cite{Jung_2018_ECCV}. 
}

In the testing phase, we train an online classifier using samples extracted from the first frame. For the subsequent frames, we extract proposals with the Gaussian sampling method around tracking result of the previous frame, then,  feed them into the classifier to get the response score. The proposal with the maximum score will be chosen as the tracking result of the current frame. In addition, we also use \emph{hard sample mining} and \emph{online update strategy} for better tracking.

\subsection{Implementation Details}  \label{traintestPhase}
For our baseline, we first extend MDNet/RT-MDNet into dual-modality and train it on VisEvent dataset for 50 epochs. The learning rate is 0.0001, batch size is 8, and other parameters are default. The training costs about 3 hours. For the first frame, we extract 500 positive and 5000 negative samples and train an online classifier for 50 iterations. For other trackers, the LF and MF based trackers are trained on VisEvent with its default settings. We adopt the pre-trained models of EF based trackers for the testing. All extended trackers have been released to help researchers re-produce our experiments \footnote{\url{https://github.com/wangxiao5791509/RGB-DVS-SOT-Baselines}}.

\begin{table*}
\center
\small      
\caption{Comparison of existing event datasets for object tracking. $\#$ denotes the number of the corresponding item. } \label{benchmarkList}
\begin{tabular}{l|ccccccccccccccc}
\hline \toprule [0.5 pt]
\textbf{Datasets}    &\textbf{Year}	&\textbf{\#Videos}  &\textbf{\#Frames}  &\textbf{\#Resolution}  &\textbf{\#Attributes} &\textbf{Aim} &\textbf{Absent}  &\textbf{Color} &\textbf{Real} &\textbf{Public}  \\ 
\hline
\textbf{VOT-DVS} \cite{hu2016dvs}     &2016    &60           &-   &$240 \times 180$  	 &-   &Eval   &\xmark   &\xmark     &\xmark     &\cmark     \\
\textbf{TD-DVS} \cite{hu2016dvs}        &2016     &77          &-   &$240 \times 180$  	 &-   &Eval  &\xmark    &\xmark     &\xmark     &\cmark     \\
\textbf{Ulster} \cite{liu2016combined}   &2016      &1     &9,000  		 &$240 \times 180$  	 &-   &Eval  &\xmark  &\xmark     &\cmark     &\xmark    		\\
\textbf{EED} \cite{mitrokhin2018event} &2018     &7     &234  &$240 \times 180$  	 &-   &Eval  &\xmark   &\xmark      &\cmark     &\cmark     \\
\textbf{FE108} \cite{Zhang2021FE108}	&2021		&$108$     &208,672   &$346 \times 260$  	 &-   &Train/Eval      &\xmark     &\xmark     &\cmark   &\cmark    \\
\hline
\textbf{VisEvent (Ours)}  	&2023     &$820$       &371,127      	 &$346 \times 260$ 	&17     &Train/Eval &\cmark &\cmark          &\cmark    &\cmark     \\
\hline \toprule [0.5 pt]
\end{tabular}
\end{table*}

\section{VisEvent Benchmark Dataset}

\subsection{Protocols } 
The VisEvent is developed to provide a dedicated platform for the training and evaluation of Visible-Event tracking algorithms. Therefore, we obey the following protocols when constructing our benchmark:
\textbf{1). Large-scale:} It is important to provide a huge amount of video sequences for data-hungry deep trackers. We collect 820 video pairs with an average of $450$ frames for each video. 
\textbf{2). High-quality dense annotations:} Our dataset is densely annotated for each frame and is independently checked by a professional labeling company and two PhDs.    
\textbf{3). Short-term \& long-term tracking:} Our dataset contains 709 and 111 videos for short-term and long-term tracking which will be beneficial for constructing a robust and flexible tracker. 
\textbf{4). Long-tail distribution:} In our real world, pedestrians, and vehicles are more related to our life and the two categories occupy the majority of our videos. 
\textbf{5). The balance between dual modalities :} Since the VisEvent contains two modalities, the balance of difficult videos for each modality is very important. The cases where tracking with a single modality can already realize high performance should be avoided.  
\textbf{6). Comprehensive baselines:} We construct multiple baselines for future work to compare by extending visible trackers into their dual-modality version with various fusion strategies. Also, we propose a simple but effective cross-modality transformer based tracker as our advanced baseline approach.

\begin{figure} 
\center
\includegraphics[width=3.2in]{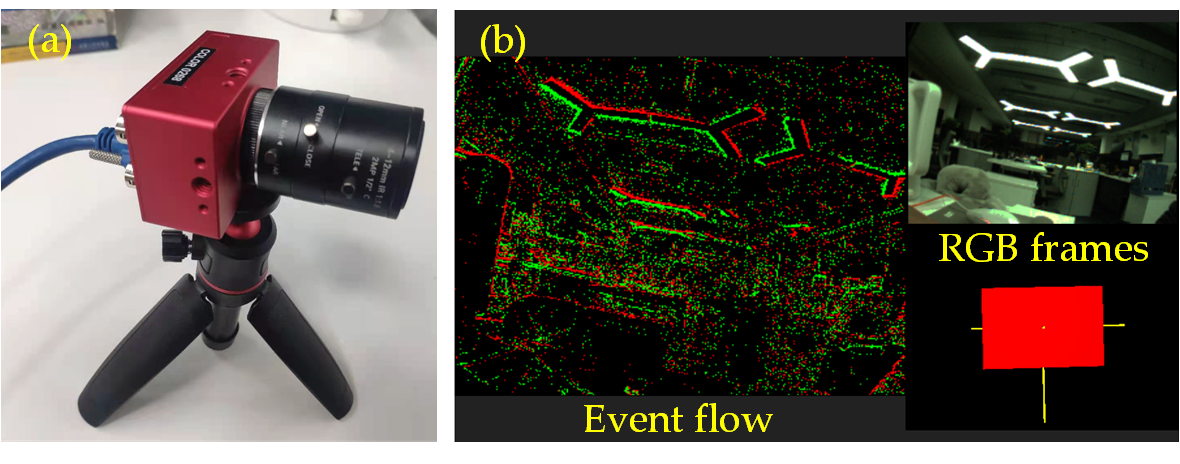}
\caption{ (a). The DVS camera used for data collection;  (b). RGB frames and event flows output from the DVS sensor. }
\label{dvs}
\end{figure}

\subsection{Data Collection and Annotation} 
\begin{figure*}[!htb]
\center
\includegraphics[width=7in]{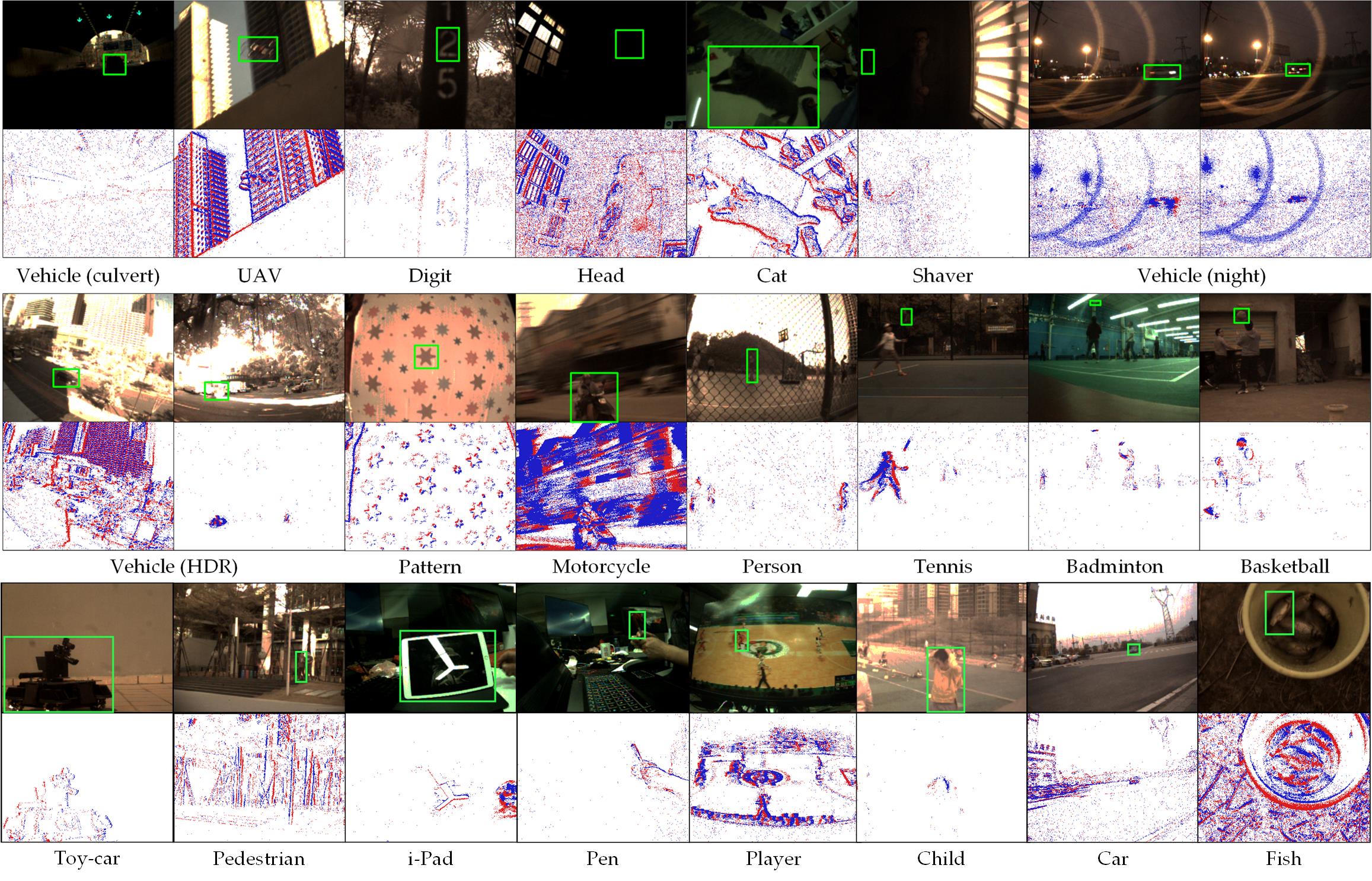}
\caption{Representative samples of our newly proposed VisEvent tracking dataset.} 
\label{videosamples}
\end{figure*}

Based on the aforementioned protocols, we first collect multiple video sequences with DVS (Dynamic Vision Sensors, as shown in Fig. \ref{dvs}), which can output visible video frames and event flows simultaneously. It is worth noting that two streams are generated from a single sensor and are already aligned by the hardware. Therefore, no external processing operations like registration in spatial and temporal views are needed. The resolution of dual modalities is $346 \times 260$. The target objects are \emph{UAV, Hand, Pen, Bottle, Tank, Toy, Car, Tennis, Pedestrian, Badminton, Basketball, Book, Plant, Shoes, Phone, Laptop, Bag} and \emph{Cat}. Parts of them are visualized in Fig.~\ref{frontimg} and Fig.~\ref{videosamples}, and more samples can be found in our demo videos.

\begin{figure}
\center
\includegraphics[width=3.2in]{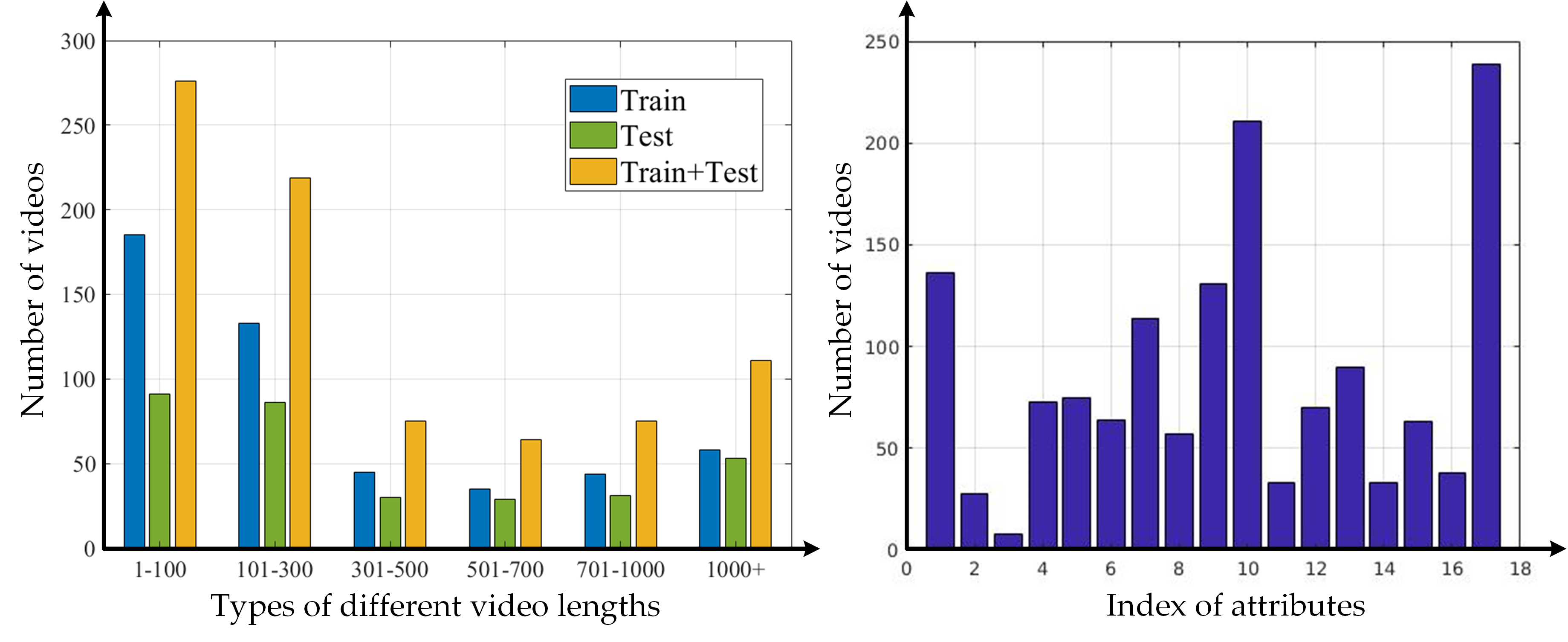}
\caption{Distribution of the proposed VisEvent dataset.}
\label{staticsvisEvent}
\end{figure}

After acquiring these videos, we first transform the output file format $*.aedat4$ into RGB and event frames $*.bmp$ and then select video clips that contain a consistent target object as one sequence. The annotation for each frame is fulfilled by a professional label company and two authors of this work checked all the annotations frame by frame. Rough annotations will be adjusted again to further ensure the accuracy of our dataset. Some samples of our dataset are visualized in Fig. \ref{videosamples}.

\subsection{Attribute Definition} 

As shown in Table \ref{AttributeList}, there is a total of 17 attributes defined in our proposed VisEvent dataset. For the RGB cameras, our dataset reflects many popular attributes in single object tracking, such as camera motion, rotation, scale variation, and occlusion. For the event cameras, since it is challenging to track the static target object, we introduce the attribute NOM (NO Motion) to evaluate the tracking performance under this situation. Other motion-related challenges are also considered, including FM (Fast Motion), MB (Motion Blur), and BOM (Background Object Motion). Our dataset also reflects the scenarios under different lighting conditions, such as LI (Low Illumination), OE (Over Exposure), and IV (Illumination Variation). It is worth noting that all our videos are with low resolution (i.e., $346 \times 260$) compared with resolution $1280 \times 720$ in GOT-10K \cite{huang2019got10k} due to the limitation of the hardware. Therefore, we do not explicitly list low-resolution attributes in Table \ref{AttributeList}. The distribution of each attribute in our dataset will be presented in subsequent sections and visualized in Fig. \ref{staticsvisEvent}.

\begin{table}
\center
\scriptsize 
\caption{Description of 17 attributes in our VisEvent dataset.} \label{AttributeList}
\begin{tabular}{l|lcccccccccccccc}
\hline \toprule [0.8 pt]
\textbf{Attributes}    &\textbf{Definition}  \\ 
\hline
\textbf{01. CM}   	    	&Abrupt motion of the camera \\	
\textbf{02. ROT}   	    &Target object rotates in the video \\	
\textbf{03. DEF}   	    &The target is deformable \\	
\textbf{04. FOC}   	    &Target is fully occluded \\
\textbf{05. LI}   	    	&Low illumination \\ 
\textbf{06. OV}   	    	&The target completely leaves the video sequence \\ 
\textbf{07. POC}   	    &Partially occluded  \\
\textbf{08. VC}   	    	&Viewpoint change  \\
\textbf{09. SV}   	    	&Scale variation  \\
\textbf{10. BC}   	    	&Background clutter  \\
\textbf{11. MB}   	    	&Motion blur  \\
\textbf{12. ARC}   	    &The ratio of bounding box aspect ratio is outside the range [0.5, 2]   \\
\textbf{13. FM}   	    	&The motion of the target is larger than the size of its bounding box  \\
\textbf{14. NMO}   	    &No motion  \\
\textbf{15. IV}				&Illumination variation  \\ 
\textbf{16. OE}			&Over exposure  \\ 
\textbf{17. BOM}         &Influence of background object motion for Event camera \\ 
\hline \toprule [0.8 pt]
\end{tabular}
\end{table}

\subsection{Statistical Analysis} 
As shown in Fig. \ref{staticsvisEvent} (left sub-figure), our proposed VisEvent tracking dataset contains 820 video sequence pairs (371,128 RGB frames total), the minimum, maximum, and average length are $18$, $6246$, and $450$ frames, respectively. The frame rate of visible videos is about 25 FPS.  
For the distribution of video length, we have $276, 222, 76, 65, 75, 111$ videos for diverse ranges, i.e., [1-100, 101-300, 301-500, 501-700, 701-1000, 1000+]. We can find that our dataset is suitable for the evaluation of both short-term and long-term tracking. 
For the challenging factors, we have [136, 28, 8, 73, 75, 64, 114, 57, 131, 211, 33, 70, 90, 33, 63, 38, 239] videos for the 17 attributes listed in Table \ref{AttributeList}, respectively. Our dataset contains many videos with camera motion, background clutter, scale variation, occlusion, and motion of distractors. Experimental results in Section \ref{experiments} show that the visual tracking problem in these scenarios is far from being solved.

\subsection{Discussion}
In this section, we give a direct comparison between Visible-Event and other dual-modal tracking tasks, including RGB-Thermal and RGB-Depth. These two tasks also attempt to fuse dual modalities for robust object tracking. For the RGB-Thermal, the thermal sensor can sense the temperature of the surface of the object and is not affected by the illumination. Therefore, it has a long sensing distance and works well in the nighttime. However, this sensor is sensitive to thermal cross-over, i.e., the image quality is bad when the target object has a similar temperature with background and motion blur. The high price is also one of the reasons restricting its wide applications. For the RGB-Depth, the depth sensors can perceive objects well in 3D space, however, it may only work well in local space due to the fact that its sensing distance is limited (usually less than 10 meters). Also, it can't handle the issue of low light and high speed. In contrast, the Event cameras, such as the Dynamic Vision Sensor (DVS) \cite{2008dvs}, are bio-inspired vision sensors that output pixel-level brightness changes instead of standard intensity frames. They offer significant advantages over standard cameras, namely a very high dynamic range, no motion blur, and latency in the order of microseconds. The following tutorials are recommended to have a general understanding of Event cameras. \footnote{\url{https://youtu.be/D6rv6q9XyWU}}

The study of object tracking using Event cameras is a new topic, therefore, many problems need to be addressed to achieve reliable object tracking in challenging scenarios. For example, how to represent the event flows to fully exploit the spatiotemporal information, and how to design efficient neural networks like spiking neural networks for effective feature learning. More importantly, there are still no public large-scale realistic visible-event datasets and baseline methods for object tracking which seriously limited the development of this research direction. In this work, we propose a large-scale benchmark dataset termed VisEvent to handle this problem, some sample images are visualized in Fig. \ref{videosamples}. In addition, we also construct multiple baseline trackers by extending visible trackers into dual-modality versions. 

\begin{figure*}[!htp]
\center
\includegraphics[width=7in]{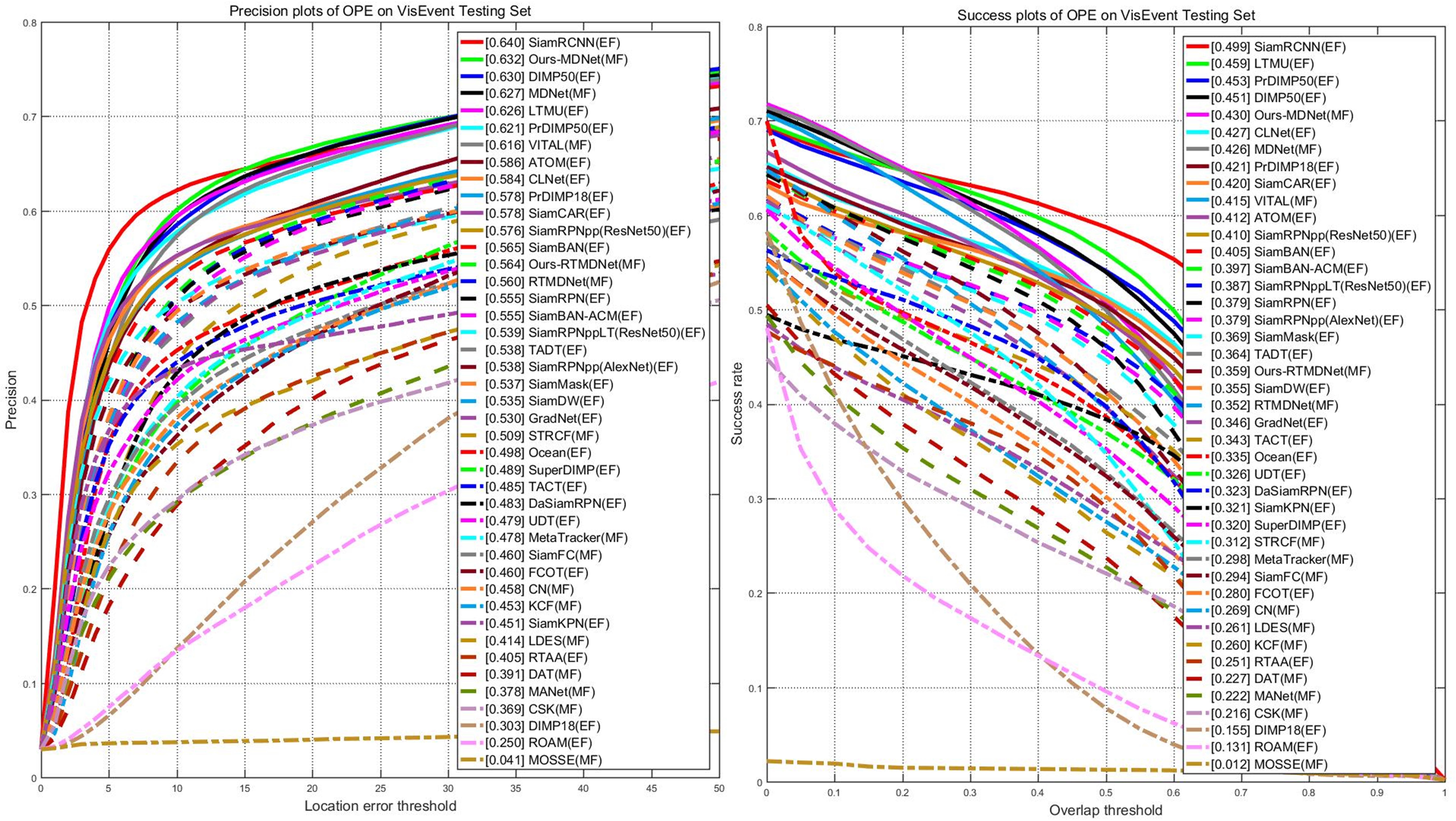}
\caption{Tracking results on the proposed VisEvent dataset (part of the constructed baselines are reported in this figure). EF and MF denote the early and middle fusion strategy used for the extension of the corresponding trackers respectively.}    
\label{benchmarkResults}
\end{figure*}

\section{Experiments} \label{experiments}

\subsection{Dataset and Evaluation Metric}    
In this work, our tracker is trained on the training subset of our proposed VisEvent dataset which contains 500 video sequences. For the testing, we evaluate and compare the trackers on the testing subset (320 videos) of  \textbf{VisEvent}, and also two simulated tracking datasets, i.e., the \textbf{OTB-DVS} and \textbf{VOT-DVS}. We adopt simulation toolkit V2E \cite{delbruck2020v2e} to complete the conversion. 
\textcolor{black}{
We also report and compare with other state-of-the-art trackers on \textbf{FE108} \footnote{\url{https://zhangjiqing.com/dataset/}}, 
and \textbf{COESOT}~\cite{tang2022coesot}\footnote{\url{https://github.com/Event-AHU/COESOT}} dataset. 
The FE108 totally contains 108 videos and the authors separate them into 76 and 32 videos for the training and testing, respectively. 
The COESOT is a newly released generic RGB-Event tracking dataset that contains 90 categories of target objects and 1354 video sequences. The training and testing subset contains 827 and 527 videos, respectively. 
}

Two popular metrics are adopted for the evaluation of tracking performance, including \textbf{Precision Plot} and \textbf{Success Plot}. Specifically, the Precision Plot illustrates the percentage of frames where the center location error between the object location and ground truth is smaller than a pre-defined threshold (default value is 20 pixels). Success Plot demonstrates the percentage of frames the IoU of the predicted and the ground truth bounding boxes is higher than a given ratio. The dataset, source code, and evaluation toolkit can be found at \url{https://github.com/wangxiao5791509/VisEvent_SOT_Benchmark}.

\subsection{Baseline Construction}  \label{baselineConst}

\textcolor{black}{To further boost the development of this research area,} in this work, we construct many baseline trackers by fusing visible frames and event flows for future works to compare.

\textbf{$\bullet$ Representative Trackers: } 
Three kinds of representative tracking frameworks are explored in this work:  \emph{1). Binary Classification based Trackers:} 
MDNet \cite{Nam2015Learning}, RT-MDNet \cite{Jung_2018_ECCV}, VITAL \cite{song2018vital}, Meta-Tracker \cite{Park_2018_ECCV}, and MANet \cite{li2019MANet}. 
\emph{2). Correlation Filter based Trackers:} 
KCF \cite{Henriques2015High},  STRCF \cite{li2018STRCF},  MOSSE \cite{bolme2010MOSSE},   CSK \cite{henriques2012CSK},   CN \cite{danelljan2014CN},  DAT \cite{possegger2015DAT},   LDES \cite{li2019LDES}.    
\emph{3). Siamese Matching based Trackers:} 
SiamFC \cite{bertinetto2016siamfc}, SiamFC++ \cite{xu2020siamfc++}, SiamRPN \cite{li2018siamRPN}, SiamRPN++ \cite{li2018siamrpn++} (AlexNet, ResNet50, and Long-term versions), ATOM \cite{danelljan2019atom}, DIMP \cite{bhat2019DiMP}, PrDIMP \cite{danelljan2020PRDiMP}, SiamRCNN \cite{voigtlaender2020siamRCNN}, Ocean \cite{zhang2020ocean}, and SiamDW \cite{siamdw2019}.

\textbf{$\bullet$ Various Fusion Strategies: }  
Three kinds of fusion strategies are considered when extending the aforementioned trackers, including: 
\textbf{1). Early Fusion} denotes the strategy that fuses the input data before feeding them into the tracking model. In this paper, two kinds of operations are explored, in more detail, we first simply \emph{add} or  \emph{concatenate} corresponding RGB and event frame as one unified data for tracking. Therefore, existing RGB trackers can be directly tested as baseline algorithms to compare for future works. 
\textbf{2). Middle Fusion} is also termed \emph{Feature Fusion} and it is widely used in current multi-modal fusion methods. In this work, we consider the following approaches for fusion. 
$(a) Concat: $ We simply concatenate the features of dual modalities to get the fused representation for tracking. 
$(b) Add: $	The two features are added together as the final features. 
$(c) 1 \times 1 ~ Conv: $ The convolutional layer with kernel $1 \times 1$  is used for fusing the feature maps. 
$(d) CAtten: $	The widely used channel attention is employed for fusion. 
$(e) SAtten: $  Spatial attention is used for feature fusion. 
$(f) CAM: $  Cross attention module proposed in \cite{suo2021CAM}. 
\textbf{3). Late Fusion}  (or response fusion) targets at combining the \emph{response score} or \emph{activation map} output from the tracking model.

\subsection{Benchmark Comparison}  

\textbf{$\bullet$ Results on VisEvent dataset.}
In this work, we construct multiple baseline methods for future works to compare on our dataset and report part of these results in Fig. \ref{benchmarkResults}. Specifically, we can find that the correlation filter based trackers achieve lower scores on this benchmark due to the manually designed features used in their model, such as HOG, gray pixels, etc. With the help of the deep features, binary classification based trackers achieve better performance than correlation filter based trackers. For example, the MDNet \cite{Nam2015Learning}, VITAL \cite{song2018vital}, RT-MDNet \cite{Jung_2018_ECCV} get $0.627|0.426$, $0.616|0.415$, $0.560|0.352$ respectively, while the CN \cite{danelljan2014CN} and KCF \cite{Henriques2015High} only achieve $0.458|0.269$ and $0.453|0.260$.

Interestingly, we also find that the Siamese network based trackers usually occupy the top few rankings of other datasets, but are not always very prominent on our dataset. Among of them, the SiamRPN++ \cite{li2018siamrpn++} achieves $0.538|0.379$, $0.576|0.410$ and $0.539|0.387$,  when AlexNet \cite{krizhevsky2012alexnet}, ResNet50 \cite{he2016deepResNet}, and long-term versions are evaluated. These results are significantly worse than MDNet and its multiple extensions which may demonstrate that the number of layers of the backbone network is not the most important role for Visible-Event tracking. The long-term version of SiamRPN++ did not increase the overall score when comparing it with its short-term version. This phenomenon fully demonstrates the challenge of our proposed tracking dataset. Compared with these works, we can attain better results with the help of the CMT module, i.e., $0.632|0.430$ on precision and success plots respectively. Our results are even better than DiMP50 \cite{bhat2019DiMP} which is a very strong tracker developed based on deep residual networks (50 layers, Ours: 3 convolutional layers). These results fully demonstrate the effectiveness and advantages of our proposed baseline tracker.

\textcolor{black}{
To compare with recent event-based trackers STNet~\cite{zhang2022STNet} and strong trackers like TransT~\cite{chen2021TransT}, Ocean~\cite{zhang2020ocean}, in this part, we evaluate our tracker on the subset of VisEvent dataset, termed VisEvent-aedat4 (i.e., the videos with aedat4 file only), by following STNet~\cite{zhang2022STNet}. As shown in Table~\ref{VisEventaedat4Results}, we can find that our MDNet-based tracker achieves $0.460|0.280$ on PR and SR metrics, which is comparable with existing strong trackers like TransT~\cite{chen2021TransT} and ATOM~\cite{danelljan2019atom}, and better than KYS~\cite{Goutam2020KYS}, DiMP~\cite{bhat2019DiMP}, etc. Our results are inferior to STNet~\cite{zhang2022STNet} which is a Swin Transformer based event tracker proposed by Zhang et al., this may be caused by the fact that the STNet considers the temporal information well using spiking neural networks. In our future works, we will consider designing advanced temporal information mining modules for high-performance tracking. 
}

\begin{table}[!htp]
\center
\scriptsize      
\caption{\textcolor{black}{Experimental results (PR$|$SR) on the VisEvent-aedat4 subset.}} 
\label{VisEventaedat4Results} 
\begin{tabular}{cccccccccc} 	 
\hline \toprule [0.5 pt]  
\textbf{STNet}~\cite{zhang2022STNet} 
&\textbf{KYS}~\cite{Goutam2020KYS} 
&\textbf{TransT} \cite{chen2021TransT} 
&\textbf{SiamRPN} \cite{li2018siamRPN}  \\ 
 $0.492|0.355$      
&$0.424|0.313$     
&$0.471|0.329$     
&$0.372|0.252$     \\  
\hline 
\textbf{ATOM} \cite{danelljan2019atom} 	
&\textbf{Ocean}~\cite{zhang2020ocean}  
&\textbf{DiMP}~\cite{bhat2019DiMP}  
&\textbf{Ours (MDNet)} \\ 
\hline 
 $0.462|0.291$  
&$0.404|0.279$      
&$0.434|0.322$       
&$0.460|0.280$       	 \\ 
\hline \toprule [0.5 pt]
\end{tabular}
\end{table}

\textbf{$\bullet$ Results on FE108 dataset.}
As shown in Table \ref{FE108Results}, we compare with multiple strong Siamese trackers on FE108 dataset, including SiamRPN++ \cite{li2018siamrpn++}, SiamBAN \cite{chen2020siamban}, SiamFC++ \cite{xu2020siamfc++}, and KYS \cite{Goutam2020KYS}. We can find that the results of these trackers on this benchmark are poor. Our tracker based on MDNet and ATOM attains $0.578|0.351$ and $0.794|0.543$, respectively, which are significantly better than these trackers. These experiments on this benchmark also validated the advantages of our trackers on RGB-DVS tracking. 

\begin{table*}[!htp]
\center
\scriptsize      
\caption{Comparison on the FE108 dataset. The results of baseline trackers are borrowed from FE108 benchmark.} \label{FE108Results} 
\begin{tabular}{cccccccccc} 	 
\hline \toprule [0.5 pt]  
&\textcolor{black}{\textbf{Zhang et al.} \cite{Zhang2021FE108}}  
&\textcolor{black}{\textbf{KYS}}~\cite{Goutam2020KYS} 
&\textcolor{black}{\textbf{CLNet} \cite{dong2020clnet}} 
&\textcolor{black}{\textbf{PrDiMP} \cite{danelljan2020PRDiMP}}  
&\textbf{SiamRPN++} \cite{li2018siamrpn++} 			
&\textbf{SiamBAN} \cite{chen2020siamban} 	 \\ 
&\textcolor{black}{$0.924|0.634$}       
&\textcolor{black}{$0.410|0.266$}       
&\textcolor{black}{$0.555|0.344$}        
&\textcolor{black}{$0.805|0.530$}       
&$0.335|0.218$  	
&$0.374|0.225$  	 \\ 
\hline 
&\textcolor{black}{\textbf{Zhu et al.}}~\cite{zhu2022GrapheventTrack} &\textbf{SiamFC++} \cite{xu2020siamfc++} 	  &\textbf{KYS} \cite{Goutam2020KYS}	  	&\textbf{ATOM} \cite{danelljan2019atom}	    &\textbf{Ours (MDNet)}  	&\textbf{Ours (ATOM)}	  \\
&\textcolor{black}{$0.859|0.549$}       &$0.391|0.238$ 		&$0.410|0.266$  		&$0.713|0.465$ 		&$0.578|0.351$   		&$0.794|0.543$ 					  		 \\
\hline \toprule [0.5 pt]
\end{tabular}
\end{table*}

\begin{table}[!htp]
\center
\scriptsize      
\caption{\textcolor{black}{Experimental results (PR$|$SR) on the COESOT dataset.}} 
\label{COESOTResults} 
\begin{tabular}{cccccccccc} 	 
\hline \toprule [0.5 pt]  
\textbf{SiamFC(MF)}~\cite{bertinetto2016siamfc}  
&\textbf{SiamBAM-ACM}~\cite{han2021SiamBANACM}  
&\textbf{SiamRPN} \cite{li2018siamRPN}  
&\textbf{RTS50} \cite{paul2022RTS}  \\ 
$0.494|0.418$      
&$0.636|0.516$     
&$0.657|0.535$     
&$0.651|0.561$     \\  
\hline 
\textbf{STARK-ST50} \cite{yan2021stark} 	 
&\textbf{Mixformer22k}~\cite{cui2022mixformer}   
&\textbf{PrDiMP18}~\cite{danelljan2020PRDiMP}   
&\textbf{Ours (MDNet)} \\  
\hline 
$0.667|0.560$  
&$0.663|0.557$       
&$0.680|0.567$       
&$0.665|0.533$       	 \\ 
\hline \toprule [0.5 pt]
\end{tabular}
\end{table}

\textbf{$\bullet$ Results on COESOT dataset.}
\textcolor{black}{
As shown in Fig.~\ref{COESOTResults}, we report our tracking results and compare them with other strong visual trackers on the recently released COESOT dataset. Obviously, our proposed MDNet-based RGB-Event tracker obtains $0.665|0.533$ on the PR and SR metrics, which is significantly better than most of the compared SOTA trackers, like the SiamBAN-ACM~\cite{han2021SiamBANACM}, SiamRPN~\cite{li2018siamRPN}, RTS50~\cite{paul2022RTS}, and comparable with STARK-ST50~\cite{yan2021stark} and Mixformer22k~\cite{cui2022mixformer}. Note that, the latter two are strong Transformer based trackers proposed in recent years. These comparisons fully demonstrate the effectiveness of our proposed cross-modality transformer fusion module for RGB-Event visual tracking. 
}

\textbf{$\bullet$ Results on Artificial OTB-DVS \cite{wu2015OTB} and VOT-DVS \cite{vot2019seventh}.}
To comprehensively validate the effectiveness of our model, we also test it on two popular tracking datasets, including OTB-DVS and VOT-DVS datasets, and report their AUC score in Table \ref{otbvotResults}. Specifically, we can find that our model achieves $0.68$ and $0.33$ based on RGB videos and Event images, respectively, on the OTB-DVS. Better results can be obtained if both of them are used, i.e., $0.69$ on this dataset. For the VOT-DVS, we get $0.39, 0.18$, and  $0.43$ on these three settings, respectively. These results fully demonstrate the effectiveness of event flows for the improvement of tracking performance. It is worth noting that we only conduct self-comparison on the two datasets and do not compare with other trackers, as the two datasets are all simulated data.

\begin{table}
\center
\small   
\caption{Tracking results of OTB-DVS, VOT-DVS, and COESOT datasets. AUC score and PR/SR are reported for the first two and third datasets, respectively.} 
\label{otbvotResults} 
\begin{tabular}{c|ccccccc} 		
\hline \toprule [0.5 pt] 
\textbf{OTB-DVS} 			&\textbf{RGB} 	  	&\textbf{Event} 	  &\textbf{Both} 	      \\ 
\textbf{Results}   	&0.68  	&0.33 	&0.69  \\ 
\hline 
\textbf{VOT-DVS} 			&\textbf{RGB} 	  	&\textbf{Event} 	  &\textbf{Both} 		  \\
\textbf{Results}    &0.39   &0.18   &0.43  \\
\hline 
\textbf{COESOT} 			&\textbf{RGB} 	  	&\textbf{Event} 	  &\textbf{Both} 		  \\
\textbf{Results} 			&0.622/0.508 	  	&0.442/0.383 	  &0.665/0.533 		  \\
\hline \toprule [0.5 pt]
\end{tabular}
\end{table}


\subsection{Ablation Study}

\noindent 
\textbf{Influence of Input Modalities:} 
In this work, we report the tracking results with a single modality to validate the effectiveness of combining visible and event sensors. As shown in Table \ref{CMAnalysis}, the baseline tracker MDNet achieves $0.605|0.412$ when only visible frames are used. If only the event frames are used, it can achieve $0.460|0.280$. It is significantly worse than tracking results with visible cameras which demonstrates that only the event sensors are not enough for practical tracking. Because it can only capture where events occurred in the scene and provide outline shape information. This information is very important for tracking but we still need the appearance and detailed texture information to discriminate the target object from other distractors. When we combine both modalities for tracking, the overall performance can be improved to $0.627|0.426$. 

In addition, we also test the PrDIMP18 tracker based on RGB and event images only and get the $0.554|0.407$ and $0.404|0.256$, respectively. When we fuse the two modalities with early fusion, we can get $0.578|0.421$, which is significantly better than a single modality only. It also attained better results on the motion blur and low illumination attributes. 
These experimental results fully demonstrate the useful clues provided by event flows. Similar conclusions can also be drawn from the results based on RT-MDNet. Therefore, it will be an interesting research direction of reliable object tracking through the collaboration of video frames and event flows.

\noindent 
\textbf{Influence of Cross-Modality Transformer:}  
To better understand the contributions of the feature fusion module in our proposed trackers, we integrate it with MDNet and RT-MDNet to check its influence on final tracking. The tracking results are reported in Table \ref{CMAnalysis}. When integrated into MDNet, clearly, its results $0.627|0.426$ can be improved to $0.632|0.430$. It also helps RT-MDNet by improving the results from $0.560|0.352$ to $0.564|0.359$. These results demonstrate the effectiveness of the feature-level information fusion module CMT for tracking.

\begin{table}
\center
\caption{Up: component analysis of our tracking model; Down: modality analysis of PrDIMP18 on all testing videos, motion blur (MB), and low illumination (LI) subset.}
\label{CMAnalysis} 
\resizebox{0.49\textwidth}{15mm}{
\begin{tabular}{c|ccccc} 		
\hline \toprule [0.5 pt] 
Index 							&Frame 	  	&Event 	 			&CMT 	  	&Ours (MDNet) 				&Ours (RT-MDNet) 			    \\
\hline 
\ding{172}   				&\cmark 		& 						&   		 		&$0.605|0.412$ 				&$0.538|0.342$   		 			 \\
\ding{173}   				& 				&\cmark 				&   		 		&$0.460|0.280$ 				&$0.380|0.216$   		 			 \\
\ding{174}   				&\cmark 		&\cmark 				&   		 		&$0.627|0.426$ 				&$0.560|0.352$   		 			 \\
\ding{175}   				&\cmark 		&\cmark 				&\cmark   	&$0.632|0.430$ 				&$0.564|0.359$   		 				 \\
\hline \toprule [0.5 pt]
Index 							&Frame 	  	&Event 	 			&PrDIMP18(ALL) 			&PrDIMP18(MB)		&PrDIMP18(LI) 	  				    \\
\hline 
\ding{176}   				&\cmark 		& 						&$0.554|0.407$ 				&$0.443|0.337$ 	 			&$0.483|0.361$   		 		 \\
\ding{177}   				& 				&\cmark 				&$0.404|0.256$ 				&$0.339|0.227$    		 	&$0.312|0.202$   		 				 \\
\ding{178}   				&\cmark 		&\cmark 				&$0.578|0.421$ 				& $0.471|0.359$   		 	&$0.517|0.375$  		 				 \\
\hline \toprule [0.5 pt]
\end{tabular} } 
\end{table}

\begin{figure}
\center
\includegraphics[width=3.5in]{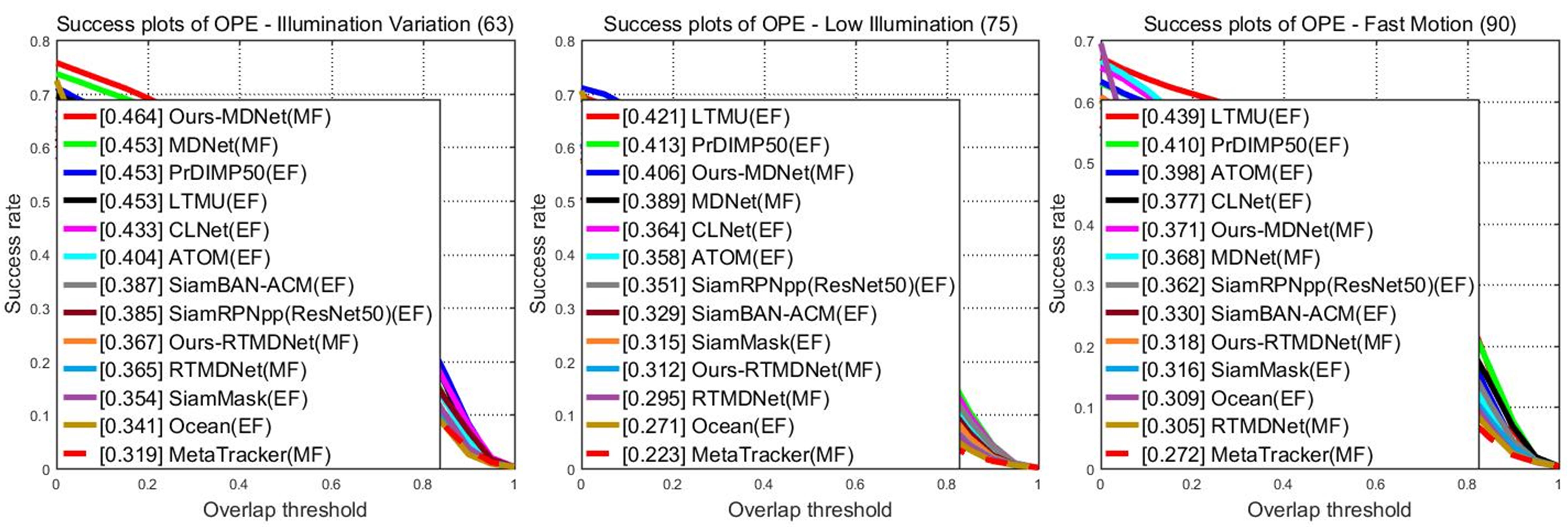}
\caption{Tracking results under different challenging scenarios. Best viewed by zooming in.}   
\label{attributeResults}
\end{figure}

\noindent 
\textbf{Influence of Challenging Factors:}
In this work, 17 attributes are considered for the VisEvent tracking dataset. In this section, we report most of them in Fig. \ref{attributeResults}, including \emph{low illumination} and \emph{fast motion}. It is easy to find that our proposed tracker attains the top-5 even the best tracking performance under these attributes. These results demonstrate the effectiveness of our proposed modules for tracking under extremely challenging scenarios. More results can be found on our project page.

\begin{figure}[!htp]
\center
\includegraphics[width=3.5in]{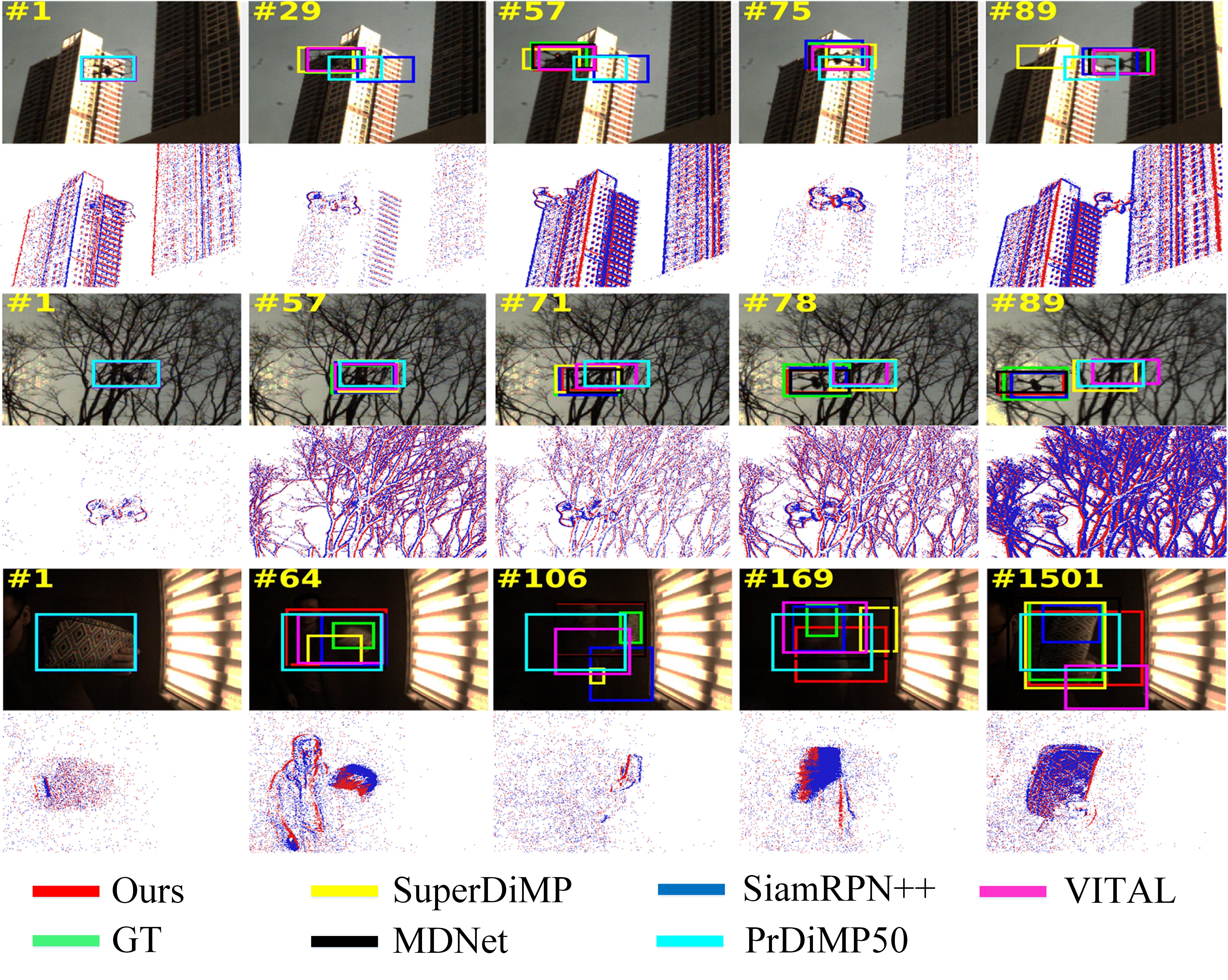}
\caption{Visualization of tracking results of our proposed tracker and other SOTA trackers on the VisEvent dataset.}    
\label{visResults}
\end{figure}

\subsection{Results of Various Fusion Strategies}   

\textcolor{black}{
In this section, we test different feature fusion strategies based on MDNet~\cite{Nam2015Learning} for RGB-Event visual tracking, as shown in Table~\ref{FusionAnalysis}. To be specific, the widely used fusion methods like \emph{concatenate}, \emph{add}, and \emph{convolution fusion with kernel size} $1 \times 1$. The \emph{channel attention (CAtt)}, \emph{spatial attention (SAtt)}, \emph{cross-attention (CAM)} used in the work~\cite{suo2021CAM}, are also exploited for multi-modal fusion. 
}
According to Table \ref{FusionAnalysis}, we can find that the simple concatenate of dual features can bring the best tracking performance, i.e., $0.627|0.426$. Interestingly, the add, channel attention, and spatial attention all achieve inferior results. We think this may be caused by the fact that the event images only provide shape information and it may hurt the visible features by simple add, and two modal information are saved to the greatest extent with the concatenating operation. 

\begin{table}[!htp]
\center
\tiny 
\caption{Results with various fusion methods.} \label{FusionAnalysis} 
\begin{tabular}{c|cccccccc}  
\hline \toprule [0.5 pt] 
Method 		&Concat  			&Add 			&$1 \times 1$ Conv 	  	&CAtt				&SAtt 		 &CAM  &CMT (Ours) \\
\hline 
Pre. Plot 		&$0.627$				&$0.471$ 	&$0.617$ 						&$0.456$   		 &$0.455$   	 		&$0.595$	   &$0.632$   \\
Suc. Plot 		&$0.426$				&$0.287$ 	&$0.422$ 						&$0.273$   		 &$0.270$   	 		&$0.402$    &$0.430$	 \\
\hline \toprule [0.5 pt]
\end{tabular}
\end{table}

\subsection{Efficiency Analysis} 
The proposed CMT is a general module for visible-event tracking, therefore, its efficiency mainly depends on the used baseline tracker. For example, when integrating our CMT into dual-modality RT-MDNet, it can run at about 14 FPS. With the help of CMT, we achieve the best tracking performance on the proposed benchmark dataset.

\subsection{Visualization} 
In addition to the aforementioned quantitative analysis, in Fig. \ref{visResults}, we also give some visualization of our tracking results and compared trackers. We can find that the current strong tracker PrDiMP50 \cite{danelljan2020PRDiMP}, SuperDiMP \cite{bhat2019DiMP}, and SiamRPN++ \cite{li2018siamrpn++} still suffer from motion blur, fast motion, and low illumination, etc. The online trackers MDNet \cite{Nam2015Learning}, VITAL \cite{song2018vital}, and ours can handle these scenarios with the help of event images.

\begin{figure} 
\center
\includegraphics[width=3.5in]{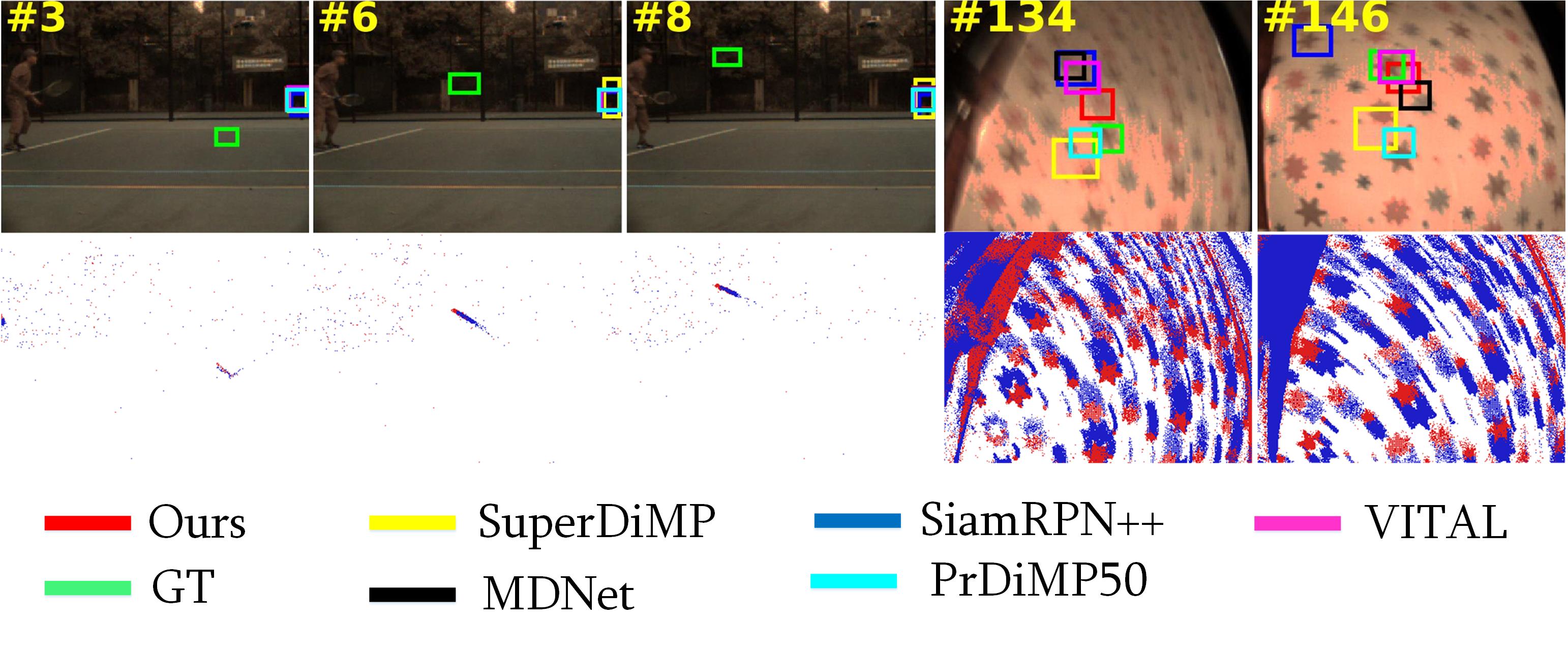}
\caption{Failed cases of our and the evaluated trackers.}   
\label{failedCases}
\end{figure}

\subsection{Failed Case Analysis}   
Although this paper achieves good results on some videos of our dataset, however, this problem is still far from being solved. The two failed cases we provided in Fig.~\ref{failedCases} are baseball and moving stars. We can find that the baseball is a fast-moving object and is almost invisible in the RGB camera. The event camera captures the moving baseball well and clearly shows the trajectory compared with the RGB frames. However, the event image representation used in this paper may be a sub-optimal choice which may lead to shape variation, as shown in the second row. For the moving stars, similar issues occurred in this case, the stacked event streams overlapped between different stars. We believe the event image representation stacked in a fixed time window is the key reason. In our future works, we will consider adopting other event representations like event points or voxels to better capture the spatiotemporal information for visual object tracking.

\section{Conclusion and Future Works}
In this paper, we propose a new and large-scale object tracking benchmark by combining the visible and event sensors. It targets providing the characteristic of high dynamic range and high temporal resolution for standard visible cameras with biologically inspired event cameras. This will widely extend the applications of current visual trackers in practical scenarios, such as high speed, low light, and cluttered backgrounds. Our dataset contains 820 video pairs that are collected with DVS cameras and it involves multiple types of objects and scenarios. We provide multiple baseline methods by extending current state-of-the-art trackers into dual-modality versions. In addition, we also designed a simple but effective baseline tracker by developing cross-modality transformer modules for interactive feature learning and fusion. Extensive experiments on the proposed VisEvent dataset fully demonstrate its good performance. We hope this work will boost the development of object tracking based on neuromorphic cameras. In our future works, we will continue to explore new architectures of pure spiking neural networks for this tracking task.


\noindent \textbf{Acknowledgement:~} This work is supported by the National Natural Science Foundation of China (No. 62102205, 62027804, 61825101), Australian Research Council Projects IH-180100002, Multi-source Cross-platform Video Analysis and Understanding for Intelligent Perception in Smart City (NO. U20B2052), Beijing Institute of Technology Research Fund Program for Young Scholars. The authors also acknowledge the High-performance Computing Platform of Anhui University for providing computing resources.

{
\bibliographystyle{IEEEtran}
\bibliography{reference}
}

\end{document}